# Physics-informed Convolutional Recurrent Surrogate Model for Reservoir Simulation with Well Controls

Jungang Chen

jungangc@tamu.edu

*Harold Vance Department of Petroleum Engineering, College of Engineering,*

*Texas A&M University, College Station, Texas, USA*

Eduardo Gildin, Ph.D.

*Professor, Harold Vance Department of Petroleum Engineering, College of Engineering,*

*Texas A&M University, College Station, Texas, USA*

and

John E. Killough, Ph.D.

*Professor(retired), Harold Vance Department of Petroleum Engineering, College of Engineering,*
*Texas A&M University, College Station, Texas, USA*

# Physics-informed Convolutional Recurrent Surrogate Model for Reservoir Simulation with Well Controls

**ABSTRACT**

This paper presents a novel surrogate model for modeling subsurface fluid flow with well controls using a physics-informed convolutional recurrent neural network (PICRNN). The model uses a convolutional long-short term memory (ConvLSTM) to capture the spatiotemporal dependencies of the state evolution dynamics in the porous flow. The ConvLSTM is linked to the state space equations, enabling the incorporation of a discrete-time sequence of well control. The model requires initial state condition and a sequence of well controls as inputs, and predicts the state variables of the system, such as pressure, as output. By minimizing the residuals of reservoir flow state-space equations, the network is trained without the need for labeled data. The model is designed to serve as a surrogate model for predicting future reservoir states based on the initial reservoir state and input engineering controls. Boundary conditions are enforced into the state-space equations so no additional loss term is needed. Three numerical cases are studied, demonstrating the model's effectiveness in predicting reservoir dynamics based on future well/system controls. The proposed model provides a new approach for efficient and accurate prediction of subsurface fluid flow, with potential applications in optimal control design for reservoir engineering.

## 1    Introduction

Surrogate modeling, also known as proxy modeling or metamodeling, refers to the construction of a computationally efficient approximation model of a complex and computationally expensive simulation model. Surrogate models have been widely used in engineering design and optimizations, such as aircraft design and optimization, weather simulations and forecast, resources management and etc. [1, 2]. In the field of reservoir simulation, it is necessary to develop surrogate models because the full order simulation models are usually computationally expensive, requiring significant computing resources and time to run. Surrogate models serve as proxies for the simulation models with much lower computational costs. These surrogate models can be constructed by training machine learning models, such as neural networks or regression models, on a set of pre-simulated data from the full simulation model. Surrogate models can then be used to make predictions of the reservoir behavior at a much lower computational cost, which enables the application of these models for tasks such as inverse modeling and optimization that would be infeasible with the full order simulation model.

Deep learning techniques can be used for surrogate modeling by training a neural network to learn the input-output relationship of a complex system. The neural network is trained using a set of input-output pairs, which can be obtained by simulating the complex system or by using historical data. [3] presents a deep-learning-based surrogate model for predicting pressure and saturation maps and well rates for new geological realizations in channelized geological models, it shows that it achieves substantial speedup when used for data assimilations. [4, 5] proposes a method for simulating subsurface flow using deep-learning-based reduced-order modeling (ROM) frameworks. The model is based on an existing embed-to-control (E2C) framework and includes an auto-encoder and a linear transition model. The ROM is trained using 300 high-fidelity simulations and is shown to achieve significant speedups in online computations under time-varying well controls compared to full-order simulations. Despite their effectiveness, deep learning-based surrogate models have some limitations. These models are solely data-driven and may not incorporate physical constraints, resulting in unrealistic predictions and limited applicability. Additionally, these models cannot fully replace high-fidelity numerical simulations since they require a large quantity of high-quality snapshots for training.

Physics-informed Neural Networks (PINNs) and Physics-informed Convolutional Neural Networks (PICNNs) are another deep learning technique that incorporate known physical laws or principles into their architecture, allowing them to learn from scarce data while satisfying the governing equations of the system. The basic idea behind PINNs is to compile the physics-based term to the loss function and take advantage of automatic differentiation to calculate derivative terms in the residual [6]. On the other hand, PICNNs [7, 8] are variants of PINN that leverages convolutional neural network to process spatial data, such as images and grids. Unlike PINN, they use convolutional filters which derives from numerical techniques (e.g. finite difference, finite volume, finite element methods) to approximate the derivative terms in the governing equations. [9, 10] presents a framework called theory-guided convolutional neural network (TgCNN) for solving porous media flow problems in oil reservoirs. The TgCNN approach involves incorporating the residuals of discretized governing equations during the training of convolutional neural networks (CNNs) to improve their accuracy compared to regular CNN models. The trained TgCNN surrogates are further used for inverse modeling by combining them with the iterative ensemble smoother algorithm, and improved inversion efficiency is achieved with sufficient estimation accuracy. Reference [11] presents a deep learning approach called Physics-Informed Deep Convolutional Neural Network (PIDCNN) for simulating and predicting Darcy flows in heterogeneous reservoir models without requiring labeled data. It shows that PIDCNN can accurately simulate transient Darcy flows in homogeneous and heterogeneous reservoirs and can be trained as a surrogate to predict the transient flow fields of reservoir models not included in training. The aforementioned approaches do not incorporate time-varying well controls as an input to the model, hence they cannot predict the behavior of reservoirs in response to future well controls.

Although PINNs and PICNNs can learn continuous time dynamics by introducing time variable or time maps as network inputs [6, 10], they do not explicitly capture the discrete time state evolution of the dynamical systems. Recurrent neural networks (Long Short-Time Memory, vanilla RNN) are designed to process sequential data and have the ability to maintain a memory of past inputs. In particular, RNNs can be viewed as discrete-time dynamical systems that generate sequences by recursively applying a nonlinear function to a hidden state vector. The hidden state vector captures information about the past inputs and is updated at each timestep. The output of the RNN at each time step can be interpreted as a prediction based on the current input and past inputs. The connection between RNNs and dynamical system control has been extensively studied. [12] proposes a recurrent neural network for modelling dynamical systems. The paper explains that the network can learn from multiple temporal patterns and can be used to model real-world processes where empirical measurements of external and state variables are obtained at discrete time points. [13] explores the use of RNNs for analyzing and forecasting spatiotemporal dynamical systems. The authors present three case studies where RNNs are used to reconstruct correct solutions for a system with a formulation error, reconstruct corrupted collective motion trajectories, and forecast streamflow time series possessing spikes.

Inheriting the strengths of both PICNN and RNNs and based on the work of [14] and [15], we presented a model that is suited for PDE governed dynamical systems. In contrast to previous approaches, our model can learn the discrete-time dynamics of state-space systems with controls by reformulating the ConvLSTM. The network takes initial reservoir state and a sequence of well controls as inputs and is trained by minimizing the residual of state-space equations of the system. The time derivative is calculated using backward Euler. This objective of this work is to predict reservoir flow quantities using image-to-image regression approach, with the input being the well controls (bottom hole pressure, injection rates) and output being state variables (pressure, saturation, etc.). Our model can also be extended to optimization problems for designing optimal well controls.

The paper is structured as follows: Firstly, it describes a state-space model for reservoir flow in porous media. Secondly, it provides a detailed description of the proposed model, with a highlight of how dynamical systems can be modeled by RNNs. Finally, the paper presents the results of three numerical case studies. These case studies include constant well control in a homogeneous oil reservoir, time-varying well control in homogeneous reservoirs, and time-varying control in heterogeneous reservoirs.

## 2    A State-Space Perspective of Reservoir Flow in Porous Media

Understanding reservoir flow equations from a system and control perspective is crucial in designing and implementing effective well control strategies that improve production efficiency and reduce operational costs in subsurface reservoirs [16]. By treating the reservoir flow as a dynamic system, we can use mathematical models to describe the behavior of fluid flow over time, and use control techniques to manipulate the system inputs (e.g. well controls) to achieve desired outputs (e.g. production rate, net present value).

### 2.1    Reservoir Flow in Porous Media

In this study, we consider a transient single-phase Darcy flow in oil reservoirs. If assuming the fluid is slightly compressible, the resulting governing equation that describes the conservation of mass reads as follows,

$$\rho_f \phi \frac{\partial P}{\partial t} + \nabla \cdot \left( \rho_f \frac{\boldsymbol{K}}{\mu} \cdot \nabla P \right) = q$$

Where $\rho_f$ represents fluid density, $\phi$ is the rock porosity, $\boldsymbol{K}$ denotes the permeability tensor, $\mu$ is the fluid viscosity and $q$ is the source/sink term. P denotes the fluid pressure. The governing equation is subject to different types of boundary and initial conditions, in our work, a no flow boundary condition is considered. Traditionally, a cell centered finite volume scheme, referred as two-point flux approximation (TPFA), is applied to discretize the porous flow partial differential equations. For source/sink terms, either flow rate or bottom hole pressure (BHP) can be prescribed. Flow rate may be incorporated directly, while BHP is prescribed indirectly using a Peaceman well model at the well block, which is expressed as:

$$q_i = PI_i \times (P_i - P_{wf})$$

Where $PI_i$ is the productivity index at the well block, $P_i$ is the pressure at the well block and $P_{wf}$ the prescribed BHP. For an isotropic reservoir where $K_x = K_y$, productivity index $PI_i$ can be calculated as follows:

$$PI_i = \frac{2\pi K_{x,i}\Delta z}{\mu \ln(\frac{r_e}{r_w} + s)}$$

Where $r_e = 0.14(\Delta x^2 + \Delta y^2)^{1/2}$ is the effective radius of the well block, $r_w$ represents the wellbore radius and $s$ denotes the skin factor of the wellbore.

## 2.2    State Space Representation

A state space equation is a mathematical representation of a system's behavior over time. It is a set of first-order differential equations that describe the evolution of the system's state variables, which represent its physical or abstract quantities, such as position, velocity, temperature, or pressure. The state space equation is usually written in the form [17]:

$$\dot{x} = f(x, u, t)$$
$$y = g(x, u, t)$$

Where $x$ is the state vector, $y$ is the output/observation vector, $u$ is the input/control vector, $t$ is time and $f$ and $g$ are nonlinear functions that describe the system's dynamics and observation equations, respectively. The solution of the state space equation gives the time-evolution of the system's states, which can be used to predict its future behavior and to design optimal control strategies.

To handle time-varying inputs/controls, a discretized state space equation is commonly used.

$$x_{k+1} = f(x_k, u_k)$$

Where $x_k$ is the state vector at time step k, $u_k$ is the control input at time step k, and $f$ is the system dynamics function that describes how the state evolves over time.

In the context of reservoir simulation, the formulation of the state space equation involves spatial discretization of the governing equations in **2.1** and incorporation of prescribed boundary conditions. After spatial discretization, the state space representation describing the reservoir flow system can be written as:

$$V\dot{x} = Tx + Bu$$

$V$ is a diagonal matrix that represents the accumulation of fluid in the reservoir, and it has a size of $n \times n$, where $n$ is the number of grid blocks in the reservoir. $T$ denotes the transmissibility matrix which is a symmetric band matrix with a dimension of $n \times n$. B is a location matrix whose elements are mostly zeros, with the non-zero terms encoding the location information of well controls, it has a dimension of $n \times m$. $x$ and $u$ are the state variable and control variable, respectively. $x = [P_1, P_2, ..., P_n]^T$, $u = [u_1, u_2, ..., u_m]^T$ with $n$ being the total number of gridblocks and $m$ the number of well controls.

## 3  Physics-informed Convolutional Recurrent Neural Network (PICRNN)

The architecture of the convolutional recurrent neural network comprises of three main components: the convolutional encoder, the convolutional LSTM layer and the convolutional decoder. The convolutional encoder compresses the input data, which includes the previous state and control, to a lower-dimensional latent space. The convolutional LSTM layer takes the compressed input and learns the temporal dynamics of the underlying state evolution processes. The convolutional decoder up-samples output from ConvLSTM layer to the dimension that is identical to input states, and it is followed by a convolution layer that scales the output. To improve the convergence rate of the network, a residual connection is created between the previous state and the current state. The unrolled version of the proposed model is shown in **Figure 1**. At each timestep, the predicted state from the previous timestep is used as the input state for the current timestep. The output hidden and cell states from the ConvLSTM layer are also passed as inputs to the next timestep. The network is tuned through incorporating state-space equation into the loss function and drive the aggregated loss of all timesteps to zero.

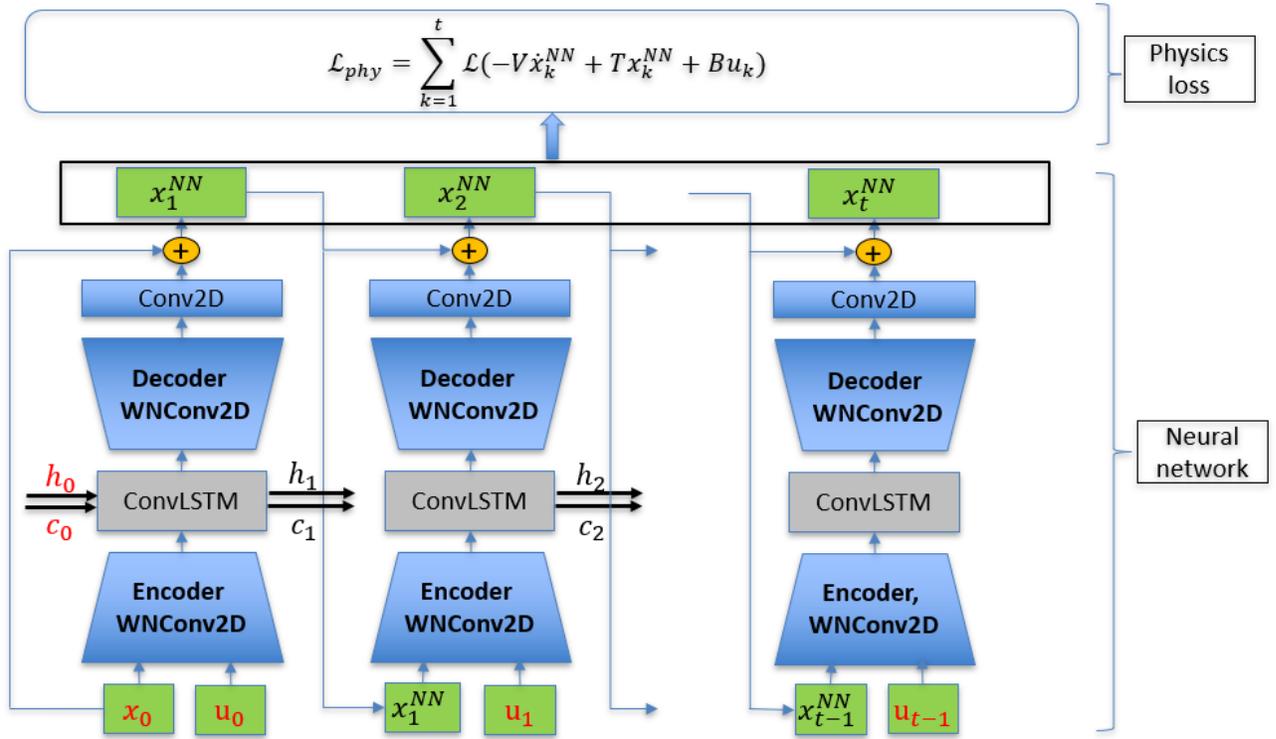

**Figure. 1.** Unrolled physics-informed convolutional recurrent network. Variables in red represent inputs of the network, including hidden and cell states of ConvLSTM layer $h_0$ and $c_0$, initial state $x_0$, and a sequence of system control $[u_0, u_1, ..., u_{t-1}]$. Predicted state $x_k^{NN}$ at timestep k serves as input for the next timestep k+1. The network weights are shared across all timesteps.

### 3.1  Convolutional LSTM

Vanilla RNN is firstly introduced to model sequential data by allowing information to flow from one step of the sequence to the next. In Vanilla RNN, the hidden state is updated at each time step using the input at that time step and the hidden state from the previous time step. A typical vanilla RNN has the following representation

$$h_{k+1} = NN(h_k, x_k, \theta)$$

Where $h_{k+1}$ is the hidden state vector, $x_k$ is the input vector and $\theta$ is the trainable weights of the network. Vanilla RNNs suffer from vanishing and exploding gradients problem, which makes it difficult to model long-term dependencies. Long Short-Term Memory (LSTM) is a type of RNN that addresses vanishing gradients of Vanilla RNN. LSTM introduces additional components called "gates" that control the flow of information through the network in order to model long-term dependencies [18]. These gates include the input gate, forget gate and output gate, which help the network selectively remember or forget information from previous time steps.

The standard LSTM architecture is designed to capture temporal dependencies in sequential data, such as text or speech. However, it does not explicitly capture spatial dependencies in the data, which can be important in many applications. Convolutional neural networks (CNNs), on the other hand, are designed to capture spatial features in image data, but they do not explicitly capture temporal dependencies. The ConvLSTM architecture combines the strengths of both LSTM and CNN architectures by incorporating convolutional layers into the standard LSTM architecture [19]. The resulting architecture can capture both spatial and temporal dependencies in the data, making it well-suited for processing spatiotemporal data. However, it cannot directly model the discrete time state-space models discussed in **2.2**. To overcome this limitation, we propose a reformulation of the Convolutional LSTM (ConvLSTM) that makes it suitable for state space models with time-varying controls. The equations for the modified ConvLSTM cell are as follows:

$$f_k = \sigma(W_{fh} * h_{k-1} + W_{fx} * x_{k-1} + W_{fu} * u_{k-1} + b_f)$$

$$i_k = \sigma(W_{ih} * h_{k-1} + W_{ix} * x_{k-1} + W_{iu} * u_{k-1} + b_i)$$

$$\tilde{C}_k = tanh(W_{ch} * h_{k-1} + W_{cx} * x_{k-1} + W_{cu} * u_{k-1} + b_c)$$

$$C_k = f_t \odot C_{k-1} + i_k \odot \tilde{C}_k$$

$$o_k = \sigma(W_{oh} * h_{k-1} + W_{ox} * x_{k-1} + W_{ou} * u_{k-1} + b_o)$$

$$h_k = o_k \odot \tanh(C_k)$$

Where $*$ represents convolutional operator, $\odot$ is element-wise product, also referred as Hadamard product. $W_{\alpha h}$, $W_{\alpha x}$, $W_{\alpha u}$ and $b_\alpha$ ($\alpha = f, i, c, o$) are trainable weights and biases. $\sigma$ denotes the sigmoid activation function layer. The ConvLSTM cell consists of four main gates: input gate $i_k$, forget gate $f_k$, output gate $o_k$, and candidate memory $\tilde{C}_k$. These gates are similar to those in a standard LSTM cell, but they are modified to include convolutional layers. In addition, there are two types of hidden states: the output state $h_k$ and the cell state $C_k$.

The forget gate decide to keep or discard the bits information of the cell state. It takes as input the previous hidden state $h_{k-1}$, the state input $x_{k-1}$, the control $u_{k-1}$, and a convolutional layer that operates on the previous hidden state. The resulting tensor is then combined with the previous hidden state and input using learned weights and biases, and passed through a sigmoid activation function to produce the forget gate $f_k$.

The input gate and candidate memory control the amount of new information that is added to the cell state. The new information is obtained by passing the previous hidden state, the input, the control and a convolutional layer that operates on the previous hidden state through a set of learned weights and biases, and a hyperbolic tangent activation function (tanh).

The output gate controls the flow of information from the cell state to the output state. It takes as input the previous hidden state $h_{k-1}$, the state input $x_{k-1}$, the control $u_{k-1}$, and a convolutional layer that operates on the previous hidden state. The resulting tensor is then combined with the previous hidden state and input using learned weights and biases, and passed through a sigmoid activation function to produce the output gate $o_k$.

The output state $h_k$ is the final output of the ConvLSTM cell. It is obtained by passing the cell state through a hyperbolic tangent activation function, and then multiplying it by the output gate activation.

The architecture of the modified ConvLSTM network is presented at **Figure. 2**.

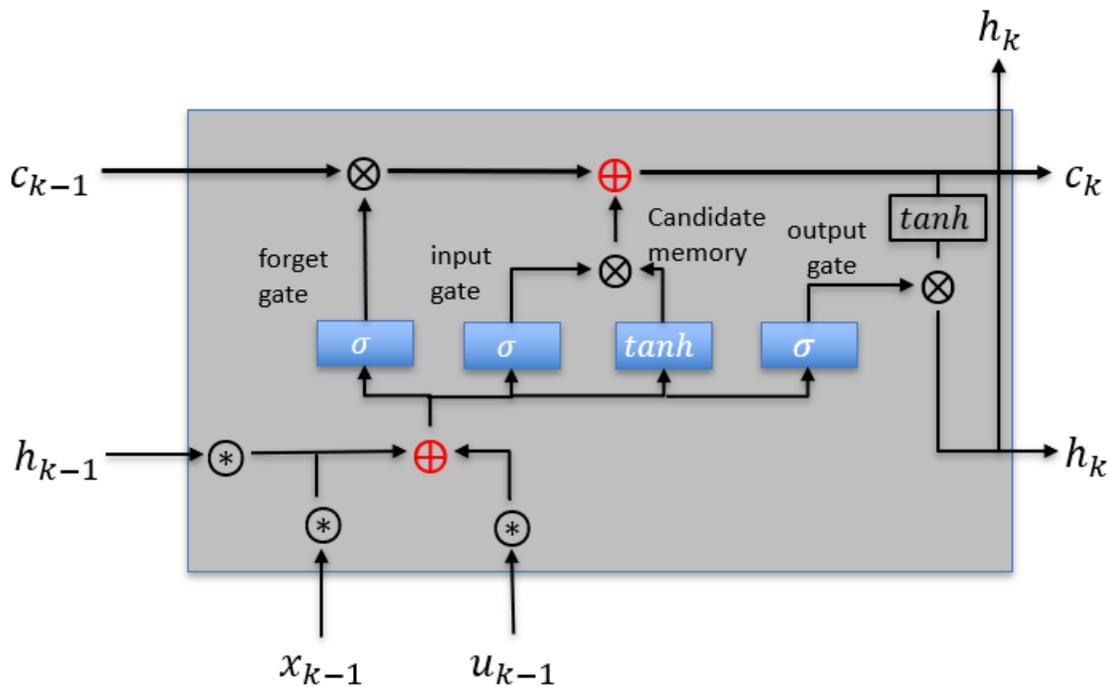

**Figure 2:** Modified ConvLSTM cell

### 3.2 Convolutional Encoder-Decoder Network

The convolutional encoder compresses the input data, which includes the previous state and control, to a lower-dimensional latent space and pass them to the following ConvLSTM layer. The encoder has two parallel structures: one for processing the input state and the other for processing the controls., as shown in **Figure 3**. For the input state encoder, it has 3 convolutional layers, with each followed by weight normalization layer (WNConv2D) and a hyperbolic-tangent activation function. The first layer has 16 channels/filters with a $4 \times 4$ kernel and 2 strides. The second and third layers have 32 and 64 filters, respectively, with the same kernel and strides as the first layer. The output tensor from this encoder is with dimension of $1 \times 64 \times 8 \times 8$. The control encoder reshapes the control vector into a 2D tensor by projecting the control vectors into corresponding well locations. The following layer downscales the 2D tensor using a reverse pixel shuffle with a downscale factor of 8, resulting in a $1 \times 64 \times 8 \times 8$ dimension tensor. The last layer of this encoder is a weight normalized convolutional layer which uses 64 filters, a $5 \times 5$ kernel and 1 stride. The output tensor from this encoder is with dimension of $1 \times 64 \times 8 \times 8$.

The architecture of the decoder is presented in **Figure 4.** The purpose of the decoder in the proposed model is to generate the state variables for the next time step from the output of the ConvLSTM layer. This can be represented mathematically as $x_k = NN(h_k)$, where $h_k$ is the output of the ConvLSTM layer. The decoder consists of 3 layers and an additional convolutional layer for scaling. For each of the first 3 layers, it is a stack of an upsampling layer with scale factor of 2, a weight normalized convolutional layer (WNConv2D) and a hyperbolic-tangent activation function. The three WNConv2D layers have 64, 32 and 16 channels, respectively and the output tensor has a dimension of $1 \times 16 \times 64 \times 64$. An additional convolutional layer, which has 1 filter, linearly scales the output tensor and reconstruct the state variable of the next timestep.

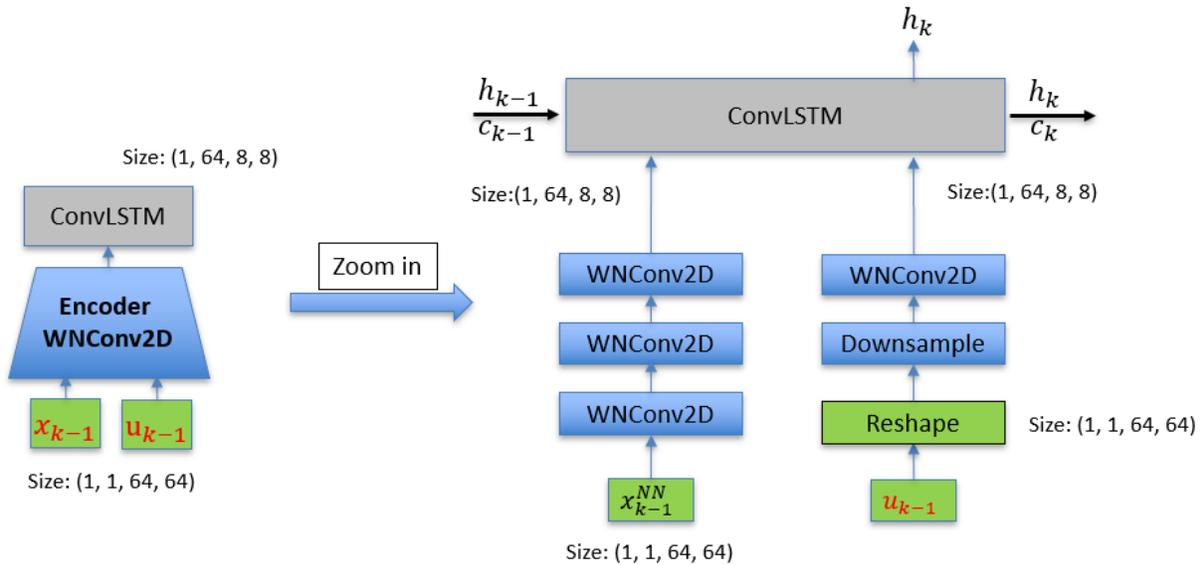

**Figure 3:** Convolutional encoder block at timestep k-1

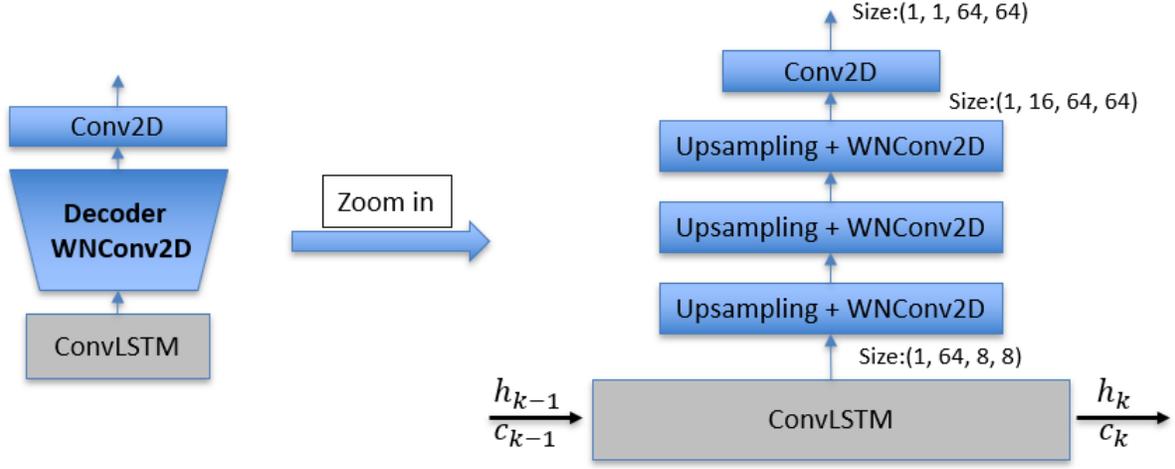

**Figure 4:** Convolutional decoder block at timestep k-1

### 3.3 Physics-informed Loss and Training

The learning process is unsupervised since no labeled data is required. The network is trained by minimizing the aggregated physics loss of all timesteps, which is expressed as:

$$\mathcal{L}_{phy} = \sum_{k=1}^{t} \mathcal{L}_s(-V\dot{x}_k^{NN} + Tx_k^{NN} + Bu_k) = \sum_{k=1}^{t} \mathcal{L}_s(-V\frac{x_k^{NN} - x_{k-1}^{NN}}{\Delta t} + Tx_k^{NN} + Bu_k)$$

$\mathcal{L}_s$ represents a smooth $L_1$ loss function operator. $x_k^{NN}$ is the network output at timestep $k-1$. $\dot{x}_k^{NN}$ represents the time derivative of output. The $\mathcal{L}_s$ has the following formulation:

$$\mathcal{L}_s = \begin{cases} \dfrac{0.5(y_{pred} - y_{true})^2}{beta}, & if\, |y_{pred} - y_{true}| < beta \\ |y_{pred} - y_{true}| - 0.5 \times beta, & otherwise \end{cases}$$

Where $beta$ is a hyperparameter, it is a fixed threshold that divides $\mathcal{L}_s$ loss into $\mathcal{L}_1$ and $\mathcal{L}_2$ regions. In our work, $beta$ is set to be 50. The $\mathcal{L}_s$ function behaves like the $L_1$ loss function when the absolute difference between the predicted and true values is small, and like the $L_2$ loss function when the absolute difference is large. This makes it more robust to outliers (e.g. well locations), as it penalizes large errors more than the $L_1$ loss function, but is less sensitive to outliers than the $L_2$ loss function.

The network parameters of the proposed model are initialized using Kaiming initialization. To train the model, Adam optimizer is used for 30000 epochs. The initial learning rate is set to 0.0023 and is decayed step-wise with a parameter of 0.995 for every 100 epochs. The experiments are conducted using a time interval of $\Delta t = 0.5$ days and an unrolled timestep of 300 with 30000 training epochs.

| ALGORITHM 1: TRAINING PHYSICS-INFORMED CONVOLUTIONAL RECURRENT NEURAL NETWORK |
|---|
| **Input:** Neural network model: $NN(\ldots, \theta)$; initialization of weights $\theta$; time interval $\Delta t$; initial state $x_0$, a time sequence of well controls $U = \{u_0, u_1, \ldots, u_{t-1}\}$, total unfolded training timesteps $t = T/\Delta t$, initial hidden states $H = \{h_0, c_0\}$, learning rate $\eta$; number of epochs $N_{epoch}$; <br> **for** epoch =1 **to** $N_{epoch}$ **do** <br>     $X \leftarrow NN(H, U, x_0, \theta)$;    # forward pass of the model, $X = \{x_1^{NN}, x_2^{NN}, \ldots, x_t^{NN}\}$ <br>     $\mathcal{L}_{phy} = 0$;    # initialize loss <br>     **for** k=1 **to** t **do** <br>        $\mathcal{L}_{phy} = \mathcal{L}_{phy} + \mathcal{L}_s(-V\frac{x_k^{NN}-x_{k-1}^{NN}}{\Delta t} + Tx_k^{NN} + Bu_k)$    # calculate aggregated loss <br>     $\nabla\theta \leftarrow Backprop(\mathcal{L}_{phy})$ ; <br>     $\theta \leftarrow optimizer.step()$;    # update network weights using predefined optimizer <br> **end** <br> **Output:** trained PICRNN model $NN(\ldots, \theta)$ |

Due to the network's recursive nature, where the output of the current time step is used as input for the next step, the training process is slow. The total training time for the homogenous reservoir cases was 36 hours, while the heterogeneous reservoir required approximately 60 hours. All three experiments were conducted using a NVIDIA A100 GPU provided by Texas A&M High Performance Research Computing.

## 4      Case studies

In the following three numerical cases, we choose the time interval $\Delta t = 0.5$ days. However, it is important to note that the time interval must be chosen carefully, as larger time steps may result in larger truncation errors in the loss function. The reference pressure results were obtained from an in-house simulator developed in previous work referenced as [20].

For these cases, a Cartesian grid with dimensions of 64 by 64 was used to discretize the reservoir domain, which has a total dimension of 40m × 40m. Two production wells were located at coordinates (10, 10) and (54, 54), as shown in Figure 5. The well radius was set to 0.09m, and the skin factor was set to 0. All other relevant fluid and rock properties are listed in Table 1.

The PICRNN network is trained using a time interval of 0.5 days and unrolled to 300 timesteps (150 days). After training, the trained network is used for prediction of reservoir states for an additional 100 timesteps (50 days) with future well controls. To perform the extrapolation, the last hidden states $H = \{h_t, c_t\}$ and the last output $x_t^{NN}$ of the trained network, in addition to future well controls, are used as inputs for the network. The network's extrapolatablity comes from its ability to learn the discrete time dynamics, which is discussed in **3.1**.

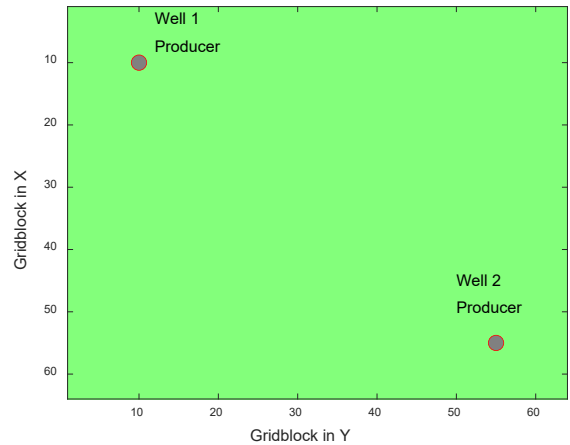

**Figure 5:** Production well locations

**Table 1:** Physical properties of the reservoir

| Rock properties | | Fluid properties | |
|---|---|---|---|
| porosity | 0.20 | Viscosity, cp | 1.13 |
| Initial pressure, psia | 3000 | Fluid Compressibility, $psia^{-1}$ | $1.0e^5$ |

### 4.1 Homogeneous reservoir with constant well control

The first case study examines a homogeneous reservoir with a permeability of 50 mD and constant well controls. Both wells have bottomhole pressures (BHPs) set to 1800 psia and the simulation lasts for 200 days with a time interval of 0.5 days.

Table 2 presents the pressure snapshots predicted by the PICRNN model, along with the corresponding reference pressure profiles predicted by the numerical FV simulator and the relative error maps. The relative error map represents the relative difference between the reference pressure and the pressure predicted by the PICRNN model across the entire reservoir domain. The snapshots were chosen at five representative time intervals during training and extrapolation, namely T=20, 60, 100 and T=160, 180 days. The results demonstrate that the pressure maps predicted by the PICRNN model are in decent agreement with the reference ones from the FV simulator, both during training and extrapolation. Moreover, the relative error maps show that most of the mismatches are concentrated around the well sink regions, with errors ranging from negligible (around 1% to 2%) to slightly larger (up to 6% in the near-well regions at T=20 days).

Overall, the results indicate that the pressure values predicted by PICRNN, both during training and extrapolation, are in excellent agreement with the reference ones obtained from the FV simulator.

**Table 2:** Comparison between predicted pressure from PICRNN against reference pressure at different time T=20, 60, 100, 160, 180 days.

| Time | | Predicted pressure $P^{NN}$ (PICRNN) | Reference pressure $P^f$ (FV simulator) | Reletive error $\|P^{NN} - P^f\|/P^f$ |
|---|---|---|---|---|
| training | T=20 days | 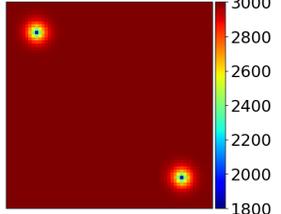 | 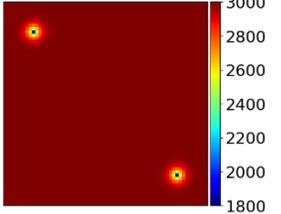 | 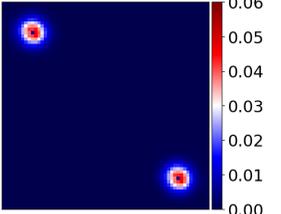 |
| training | T=60 days | 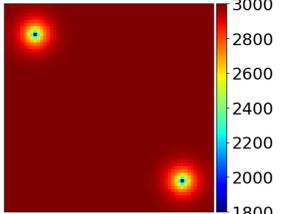 | 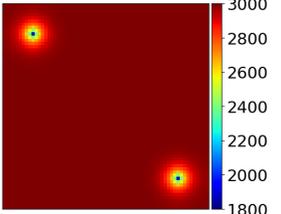 | 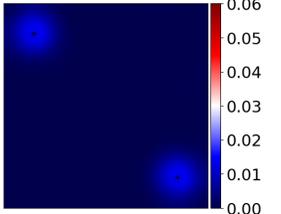 |
| training | T=100 days | 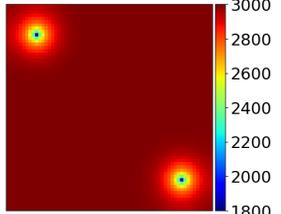 | 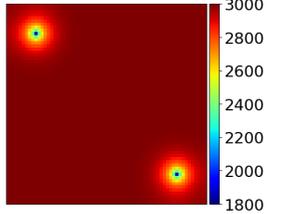 | 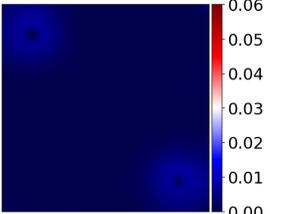 |
| extrapolation | T=160days | 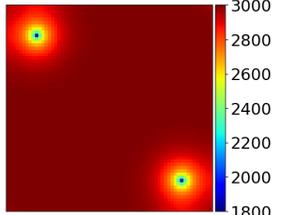 | 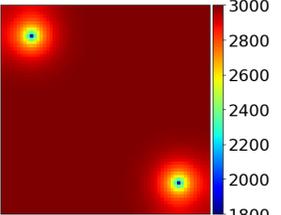 | 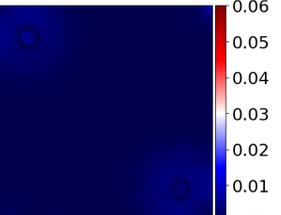 |
| extrapolation | T=180days | 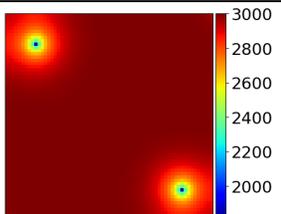 | 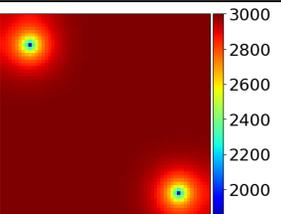 | 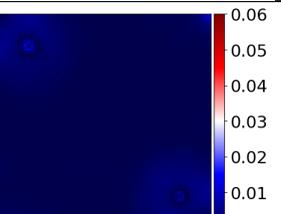 |

### 4.2 Homogeneous reservoir with time-varying well control

In this numerical case, the bottomhole pressure (BHP) of both wells is varied over time with a schedule that changes every 50 days, as shown in Figure 6. The permeability of the reservoir is set to be 50 mD, and all other conditions are the same as in case 1.

Table 3 presents the results of the PICRNN model's predictions for pressure snapshots in a homogeneous reservoir under varying well control scenarios, along with the corresponding reference pressure profiles obtained from a numerical FV simulator and relative error maps. The results indicate that the pressure maps predicted by the PICRNN model closely match the reference pressure profiles from the FV simulator, both during training and extrapolation. The relative error maps suggest that errors around the well locations range from 2% to 3%, showing that the model has good predictability and extrapolability even with different well controls in the extrapolation phase where the model has not been trained on.

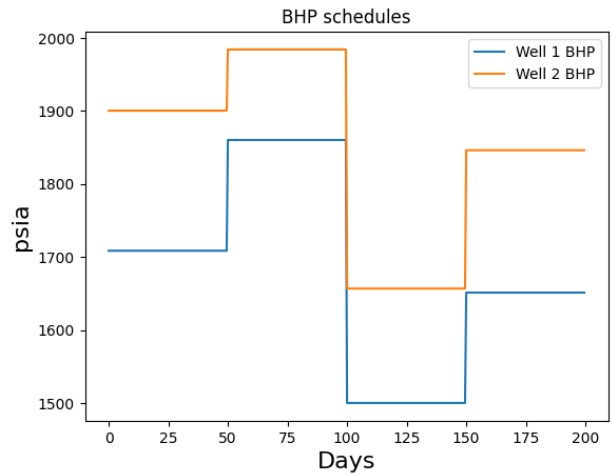

**Figure 6:** BHP schedule of both wells

**Table 3:** Comparison between predicted pressure with PICRNN and reference pressure at different timesteps T=20, 60, 100, 160, 180 days.

| Time | Predicted pressure $P^{NN}$ (PICRNN) | Reference pressure $P^f$ (FV simulator) | Reletive error $\|P^{NN} - P^f\|/P^f$ |
|---|---|---|---|
| T=20 days | | | |
14

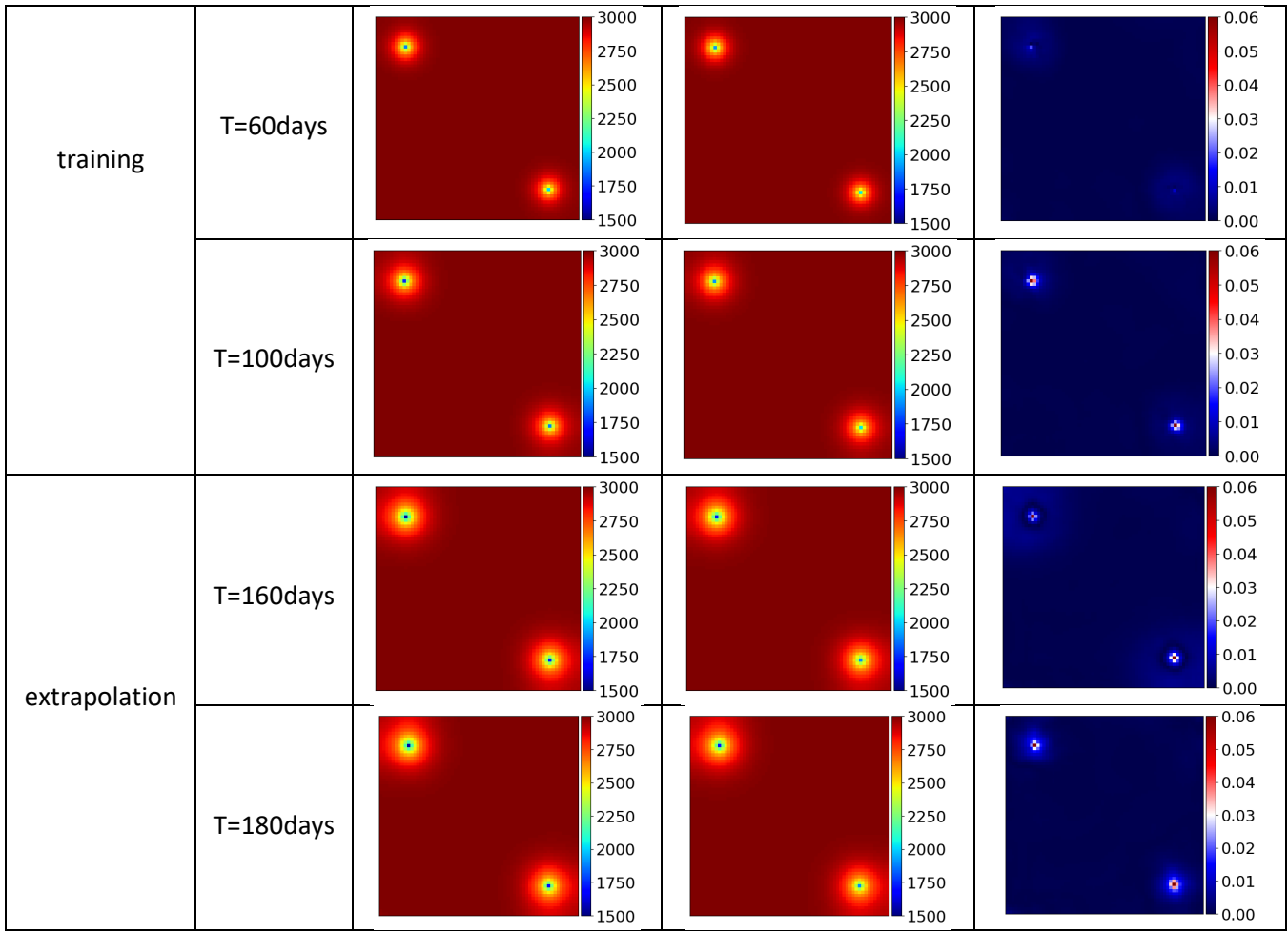

### 4.3  Heterogeneous reservoir with time-varying well controls

For this heterogeneous case, the permeability map can be found in Figure 7, and a time-varying well control strategy is applied to the production wells.

Table 4 presents the results of the PICRNN model's predictions for pressure snapshots in a heterogeneous reservoir under varying well control scenarios, along with the corresponding reference pressure profiles obtained from a numerical FV simulator and relative error maps. The results demonstrate that the pressure maps predicted by the PICRNN model are in satisfactory agreement with the reference ones from the FV simulator, both during training and extrapolation. However, compared to the results in Table 2 and Table 3, the error maps suggest slightly higher errors. This could be explained by the fact that the heterogeneous case is more challenging to train, as illustrated in the loss history in Figure 8.

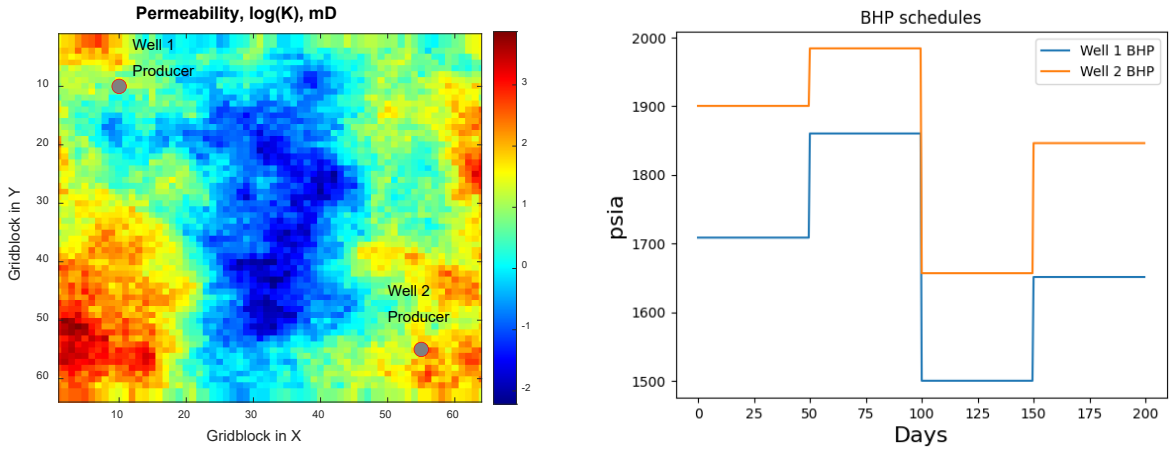

**Figure 7:** Permeability map of reservoir (left);     BHP schedule of two wells (right)

**Table 4:** Comparison between predicted pressure with PICRNN and reference pressure at different timesteps T=20, 60, 100, 160, 180 days.

| | Time | Predicted pressure $P^{NN}$ (PICRNN) | Reference pressure $P^f$ (FV simulator) | Reletive error $|P^{NN} - P^f|/P^f$ |
|---|---|---|---|---|
| training | T=20 days | | | |
| | T=60days | | | |
| | T=100days | | | |

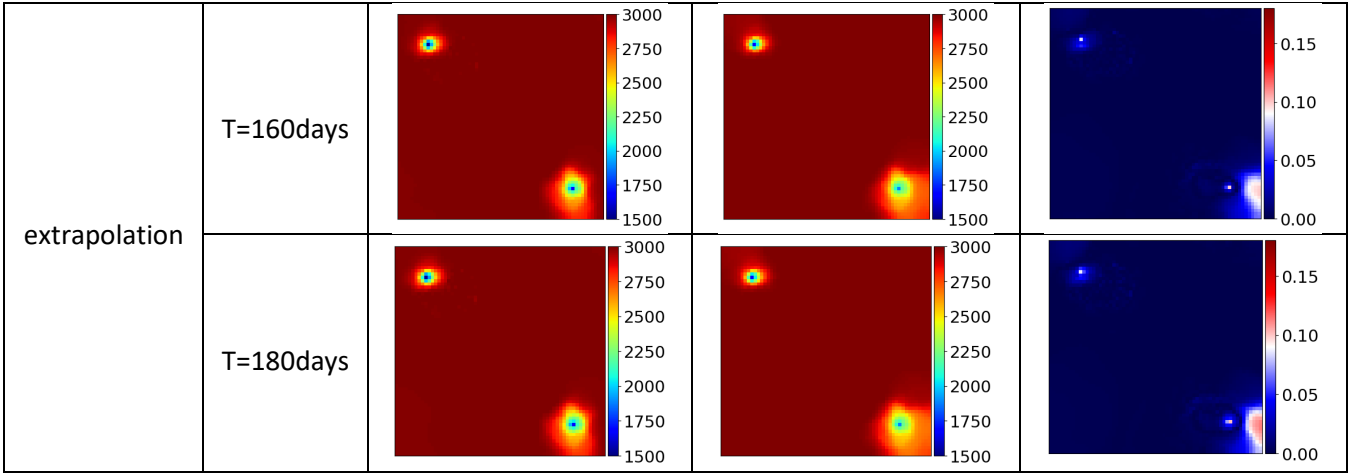

| | | | | |
|---|---|---|---|---|
| extrapolation | T=160days | | | |
| | T=180days | | | |

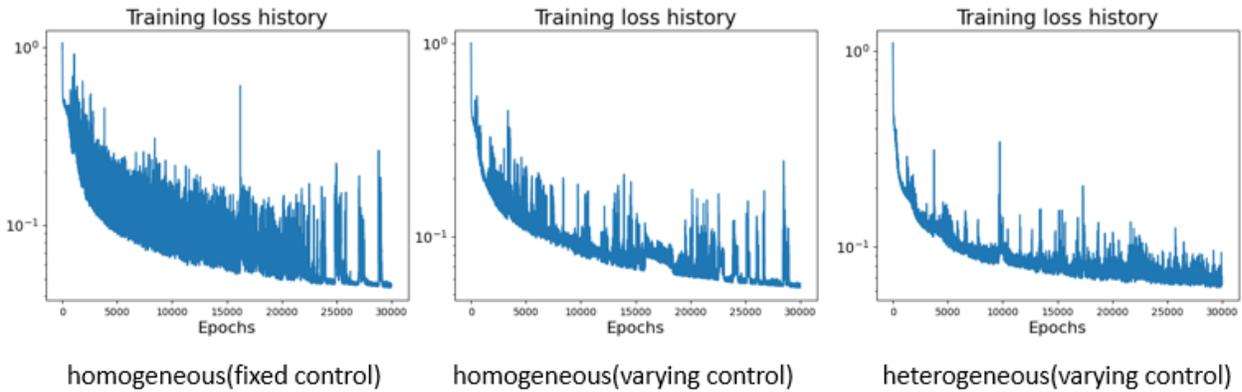

**Figure 8:** training loss history for three cases

## 5    Discussions

In this work, a physics informed convolutional recurrent neural network (PICRNN) is proposed as a surrogate model for predicting reservoir dynamics with future well controls. The network is trained to learn the underlying physics laws without labeled data. Unlike many previous parameter-to-state regression models [9, 11, 21], our model establishes a control-to-state mapping which provides reliable prediction with future well controls that has not been trained in the network. The modified ConvLSTM successfully captures the discrete time state evolution of 2D reservoirs with input controls and therefore provide decent extrapolation performance.

This study explored three different reservoir simulation scenarios: constant well control in a homogeneous reservoir, time-varying well control in a homogeneous reservoir, and time-varying well control in a heterogeneous reservoir. The proposed model showed a decent performance in predicting reservoir states, achieving less than 1% relative error in areas far from sink/source terms and approximately 3% in areas near the wells. However, the model had slightly larger errors in some areas of the heterogeneous case, possibly due to the increased difficulty of training and convergence.

It should be highlighted that the recursive nature of the network makes it challenging to train. Therefore, future studies will concentrate on developing a more efficient network and extending the analysis to more complex cases, such as incorporating well observation data to provide additional regularization for training the network, well control for two-phase reservoir flow, etc.

**ACKNOWLEDGEMENTS**

Portions of this work were conducted with the advanced computing resources provided by Texas A&M High Performance Research Computing.

**DATA AVAILABILITY STATEMENTS**

The datasets and code will be available at https://github.com/jungangc/PICRNN-well-control once published.

**DECLARATIONS**

The authors have no relevant financial or non-financial interests to disclose.